\documentclass[runningheads]{llncs}

\usepackage[year=2026,ID=6]{accv}

\usepackage{accvabbrv}
\usepackage{graphicx}
\usepackage{booktabs}
\usepackage{amsmath}
\usepackage{amssymb}
\usepackage{tcolorbox}
\tcbuselibrary{skins,breakable}
\usepackage[pagebackref,breaklinks,colorlinks,citecolor=accvblue]{hyperref}

\graphicspath{{figures/}}

\newcommand{\gemma}{Gemma~4 12B}

\newtcolorbox{findingbox}[1]{
  colback=gray!3,
  colframe=gray!50,
  boxrule=1pt,
  arc=2mm,
  left=4pt,
  right=4pt,
  top=4pt,
  bottom=4pt,
  coltitle=black,
  fonttitle=\bfseries,
  title=\textbf{#1}
}

\begin{document}

\title{Virtual Encoders in Multimodal Transformers}
\titlerunning{Virtual Encoders in Multimodal Transformers}
\author{Katsuya Ogata\inst{1} \and
Yuta Nakashima\inst{1}}

\authorrunning{K.~Ogata and Y.~Nakashima}

\institute{The University of Osaka\\
\email{\{ogata, n-yuta\}@im.sanken.osaka-u.ac.jp}}

\maketitle

\begin{abstract}
Multimodal language models traditionally rely on dedicated perceptual encoders to construct task-usable representations. More integrated architectures have recently emerged, which instead expose the multimodal transformer to lightly projected patches, audio frames, or discrete visual tokens. How does this encoding happen when such representations are not provided? We find that the transformer can internalize this missing computation, constructing task-usable perceptual representations within its own early-to-middle layers before the downstream language model. We call this computational structure a \emph{Virtual Encoder}. Across linear probing, similarities to perceptual encoders, and causal analyses, we identify signatures of this structure in models that receive perceptual tokens without encoder-derived features.
These analyses also suggest that the boundary between perception and language processing need not coincide within an architectural module. Instead, encoder-like computation can emerge as a functional regime within a multimodal transformer, providing a new perspective for understanding where and how multimodal models process perception.

\end{abstract}

\begin{figure}[t]
\centering
\includegraphics[width=\linewidth]{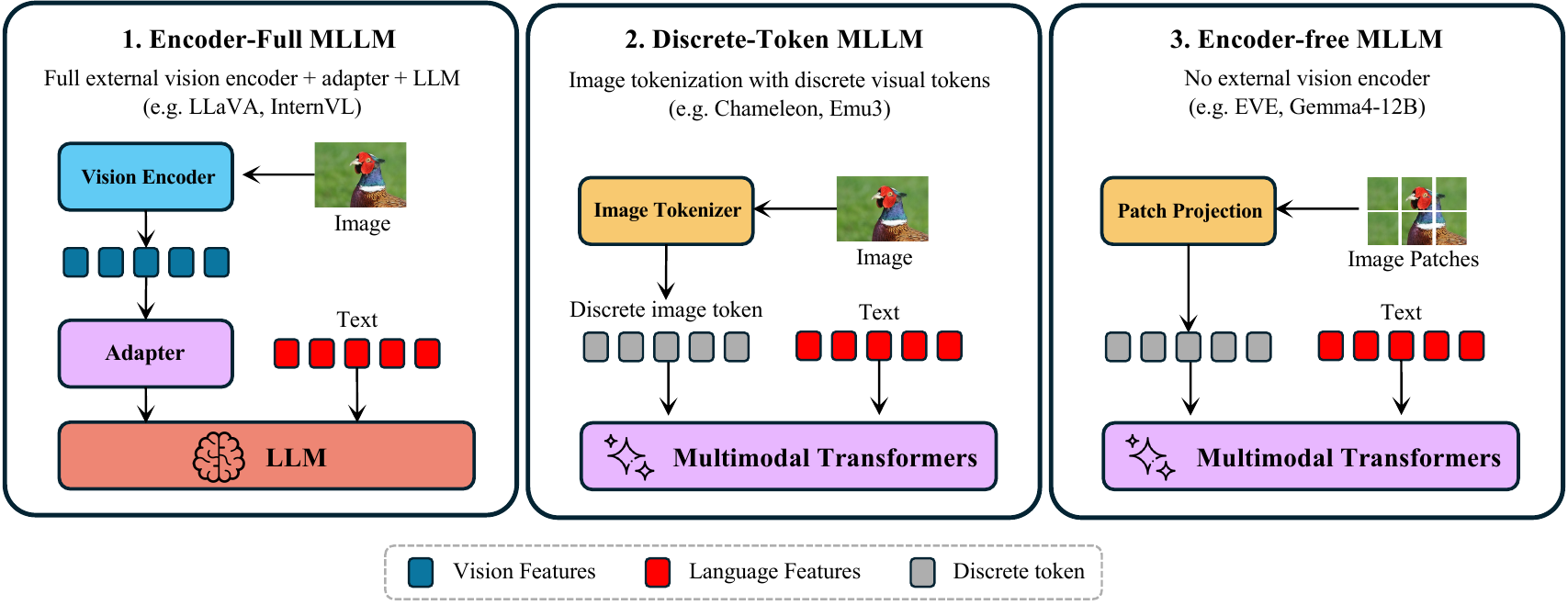}
\caption{Three MLLM families distinguished by the representation supplied to the multimodal Transformer. Encoder-full MLLMs receive continuous features from a dedicated perceptual encoder. Discrete-token MLLMs receive discrete perceptual codes, while encoder-free MLLMs receive lightly projected patches or frames. We ask whether the latter two families form a \emph{Virtual Encoder} inside the Transformer.}
\label{fig:architecture}
\end{figure}

\section{Introduction}

The development of multimodal large language models (MLLMs) has been progressively bringing
perception and language processing closer together. We organize current designs into three architectural families (Figure~\ref{fig:architecture}): \emph{Encoder-full MLLMs} use a dedicated vision or audio encoder to construct continuous perceptual features before passing them to a language model through cross-attention, a resampler, or a projector~\cite{alayrac2022flamingo,li2023blip2,liu2023llava,dai2023instructblip,bai2023qwenvl}. \emph{Discrete-token MLLMs} represent perceptual inputs as discrete codes in the same autoregressive stream as text, as in Chameleon and Emu3~\cite{team2024chameleon,wang2024emu3}. \emph{Encoder-free MLLMs} omit a dedicated perceptual encoder and feed lightly projected image patches or audio frames into the multimodal Transformer, as in Fuyu, EVE, SOLO, and Gemma~4~\cite{bavishi2023fuyu,diao2024eve,chen2024solo,gemma4team2025}. Unlike encoder-full MLLMs, these two families thus shift multimodal perception from a dedicated external encoder into a part of the transformer-based language model itself. 

This raises an intriguing question: \emph{How do perception and language processing coexist in a single Transformer-based model?} Perceptual encoding and language processing may require qualitatively different computations. Moreover, although encoder-free and discrete-token MLLMs allow perceptual and language tokens to interact from the earliest layers, meaningful cross-modal interaction may require higher-level representations to be established first. That is, architectural unification does not necessarily imply computational unification. Perceptual encoding may be organized as a distinct functional stage within the multimodal Transformer. Although architectural boundaries disappear, deeper analysis of such a model may identify the functional boundaries that replace perceptual encoders.

\begin{figure}[t]
\centering
\includegraphics[width= \linewidth]{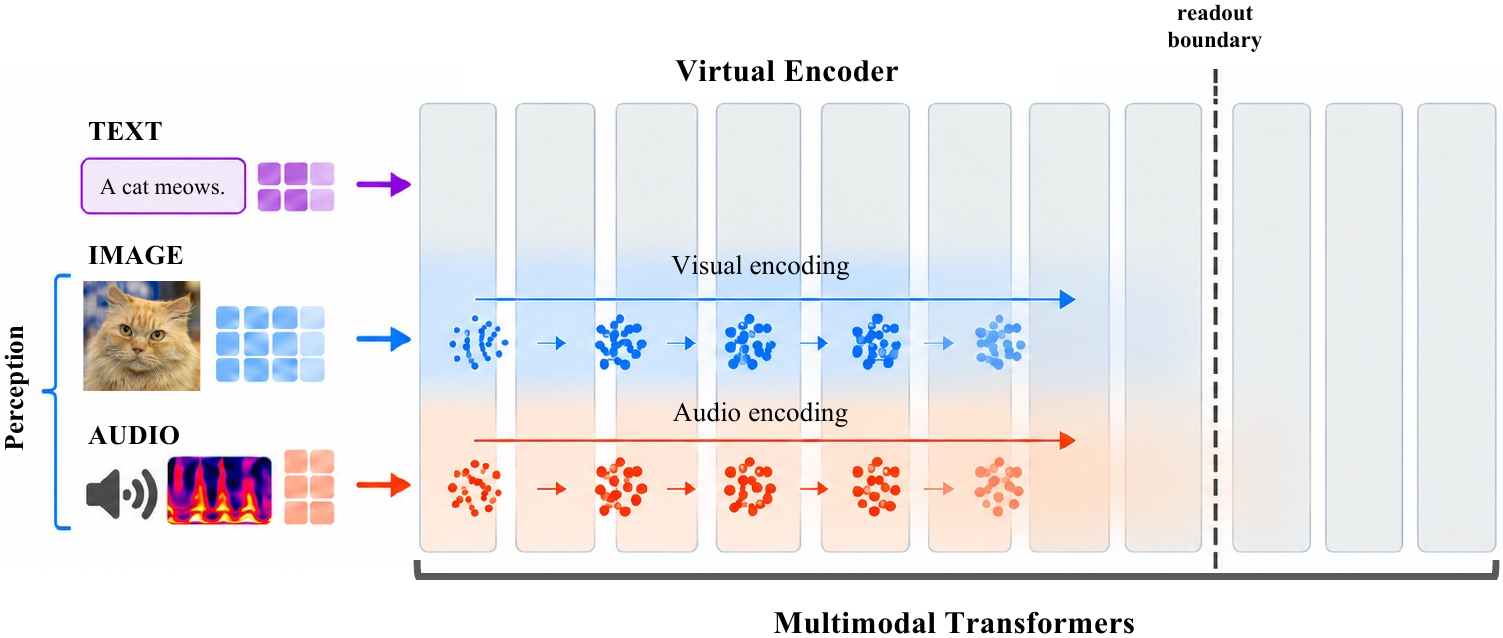}
\caption{When a multimodal transformer receives perceptual tokens
that have not already been organized by a continuous modality encoder, it must
construct a useful representation internally. In \gemma{}, an early band of
layers, which we call the Virtual Encoder, performs this encoding for both
modalities up to a mid-depth readout boundary. There the vision
representation aligns with the model's language subspace, whereas the audio
representation, though formed just as early, is not maintained downstream.}
\label{fig:teaser}
\end{figure}

We hypothesize that discrete-token and encoder-free MLLMs \emph{internalize} part of the perceptual encoding normally performed by a dedicated encoder. Specifically, an early band of Transformer layers organizes perceptual tokens into a task-usable representation and thereby assumes an encoder-like computational role. We call this functional regime a \emph{Virtual Encoder} (Figure~\ref{fig:teaser}).

Based on this hypothesis, we study encoder-free and discrete-token MLLMs, comparing them with encoder-full MLLMs through linear probing \cite{alain2016probes}, representational similarity \cite{kornblith2019cka}, and causal analyses with perturbations over image and audio inputs. We set \gemma{}~\cite{gemma4team2025} as our primary target model and consistently identify an encoder-like structure in it, forming the perceptual representation in early-to-middle layers, which are then read out. 

\textbf{Contributions}. We introduce the Virtual Encoder, a functional regime in which a multimodal Transformer internally constructs perceptual representations when no continuous encoder-derived features are supplied. We show four properties. (1) Encoder-free and discrete-token MLLMs develop semantic decodability progressively over depth, whereas encoder-full MLLMs start with already organized representations. (2) Their early hidden states exhibit encoder-like geometry. (3) The resulting representations are causally read out at a certain depths. (4) Their subsequent routing is modality- and architecture-dependent, with vision in \gemma{} transiently aligning with the language subspace near readout while audio follows a markedly different trajectory.

\section{Related Work}

\subsection{Multimodal Large Language Models}

Early MLLMs extended large language models with dedicated perceptual
encoders and lightweight bridging modules. CLIP~\cite{radford2021clip} and
SigLIP~\cite{zhai2023siglip} provided pretrained visual representations, while
models such as Flamingo~\cite{alayrac2022flamingo},
BLIP-2~\cite{li2023blip2}, LLaVA~\cite{liu2023llava}, and
InstructBLIP~\cite{dai2023instructblip} connected such features to language
models through cross-attention, Q-Former-style modules, or learned projectors.
More recent families, including Qwen-VL~\cite{bai2023qwenvl,wang2024qwen2vl,bai2025qwen3vl}
and InternVL~\cite{chen2024internvl}, further improve visual-language
alignment while retaining a dedicated vision encoder. Across these
architectures, perceptual representations are therefore constructed before
entering the language model stack. Audio-language models have followed a similar trajectory, with models such as Qwen2-Audio relying on a dedicated audio encoder to produce continuous acoustic representations before language-model processing \cite{chu2024qwen2audio}. More recent unified models such as Gemma 4 instead process audio frames within the multimodal Transformer, further blurring the boundary between perception and language processing \cite{gemma4team2025}.

Discrete-token MLLMs weaken the boundary between perceptual and language processing by placing discrete perceptual codes in the same autoregressive stream as text. Chameleon~\cite{team2024chameleon} and Emu3~\cite{wang2024emu3} follow this design. Encoder-free MLLMs remove the dedicated perceptual encoder more directly. Fuyu~\cite{bavishi2023fuyu} feeds linearly projected image patches into the Transformer, while EVE~\cite{diao2024eve,diao2025evev2} and SOLO~\cite{chen2024solo} train single-Transformer models without an external vision encoder. Mono-InternVL~\cite{luo2024monointernvl} incorporates visual experts within a monolithic language model, and Gemma~4~\cite{gemma4team2025} extends encoder-free processing to image and audio inputs.

Discrete-token and encoder-free MLLMs differ in their input units and parameterization, but neither supplies the multimodal Transformer with continuous features from a dedicated perceptual encoder. We ask how their multimodal Transformers form perceptual representations and when those representations become available to language processing.

Prior work has primarily evaluated these architectures through downstream performance, leaving their internal perceptual computation largely unexplored. We instead ask how perceptual representations are constructed across Transformer layers when no continuous encoder-derived representation is provided.

\subsection{Representation Analysis in Multimodal Language Models}

Our analysis relies on two complementary views of latent representations: what information they encode and how they are geometrically organized. Linear probes have been widely used to test what is linearly decodable at a given layer~\cite{alain2016probes}. Representation similarity is beneficial to find encoder-like structure in a model. Centered kernel alignment (CKA) \cite{kornblith2019cka}, singular vector canonical correlation analysis \cite{raghu2017svcca}, and projection-weighted canonical correlation analysis \cite{morcos2018insights} compare representations within and across networks. Intrinsic-dimension analyses further characterize how representation geometry changes across network depth~\cite{ansuini2019intrinsic,valeriani2023geometry}. A complementary line of work studies representations as linear subspaces, including the linear representation hypothesis~\cite{park2024linear}.
Principal angles give a classical comparison between two subspaces~\cite{bjorck1973numerical}, while cross-projected variance measures how much activity from one condition is captured by another condition's PCA subspace~\cite{elsayed2016reorganization}.

Such tools have identified some characteristic structures in multimodal models. The modality gap~\cite{liang2022mindthegap} describes the separation of image and text representations in a shared space, while Nikankin et al.~\cite{nikankin2025same} identify modality-specific circuits and late cross-modal alignment within VLMs. Jain et al.~\cite{jain2025elevating} further show that intermediate LLM states contain perceptual representations whose quality correlates with downstream performance, and Venhoff et al.~\cite{venhoff2025visual} study how representations produced by a dedicated vision encoder progressively map into the language feature space.
Closest to our readout analysis, Neo et al.~\cite{neo2025interpreting} and Kaduri et al.~\cite{kaduri2025whats} find that visual information flows from image tokens to text tokens mainly in the middle layers of encoder-full VLMs.

These studies characterize how a representation already built by a dedicated encoder is \emph{consumed}. We instead ask where it is \emph{constructed} when no such encoder is provided, how its construction relates to its readout depth, and how it is routed among modality-specific and language-aligned subspaces.

\section{Paired Image-Audio Dataset}
\label{sec:dataset}

We compare image and audio representations using concepts that occur in both modalities. Lin et al.~\cite{lin2023crossmodal} introduced this pairing to study whether multimodal learning can improve a unimodal classifier. Following their protocol, we pair ImageNet-1k images~\cite{deng2009imagenet} with ESC-50 sounds~\cite{piczak2015esc50}. We retain their $27$ concept pairs and add four unambiguous matches (cow, siren, car horn, and door knock), yielding a set $\mathcal{C}$ of 31 concepts. Each concept contains $40$ images and $40$ audio clips, resulting in 1,240 samples per modality. The shared concept labels let us compare layerwise probe and representation results across image and audio inputs. We split the samples into a training set $\mathcal{D}$ and a held-out set $\mathcal{D}_\text{H}$ with $|\mathcal{D}|:|\mathcal{D}_\text{H}|=3:1$, preserving the uniform concept distribution in both sets. Each element is a tuple $(K_\text{V},K_\text{A},c)$, where $K_m$ for $m\in\{\text{V},\text{A}\}$ denotes an image or audio sample and $c\in\mathcal{C}$ is its shared concept label. For the causal experiment (Section~\ref{sec:causal}), we randomly select four held-out samples per concept and denote this subset by $\mathcal{D}_\text{C}\subset\mathcal{D}_\text{H}$. We additionally group the concepts into animated/non-animated classes.

\section{The Virtual Encoder}
\label{sec:virtual-encoder}

Encoder-full MLLMs delegate perception to a dedicated encoder. A Vision Transformer or audio spectrogram model transforms the raw signal into a continuous semantic representation \emph{before} it reaches the multimodal Transformers. Encoder-free models omit this stage and take lightly projected patches or frames directly as input, while discrete-token MLLMs take discrete perceptual codes. Our central hypothesis is that, in the latter two cases, an early band of Transformer layers internalizes the missing computation and comes to play the functional role of a modality-specific encoder. We call this early-layer regime the Virtual Encoder.

To test this hypothesis, we ask three questions. (i) When do perceptual representations become semantically organized? (ii) Do intermediate representations exhibit geometry similar to that of a dedicated encoder? (iii) Until what depth are modality-token states causally required for the output? We investigate these questions through linear probing, representational similarity, and causal intervention. We primarily study the encoder-free MLLMs, \gemma{}, EVE~\cite{diao2025evev2}, and Fuyu~\cite{bavishi2023fuyu}; the discrete-token MLLM Chameleon~\cite{team2024chameleon}; and the encoder-full controls LLaVA-1.5~\cite{liu2023llava}, Qwen3-VL~\cite{bai2025qwen3vl}, Qwen3-Omni~\cite{xu2025qwenomni}, and Qwen2-Audio~\cite{chu2024qwen2audio}.

\subsection{Semantic Decodability Emerges with Depth}
\label{sec:probe}

We use linear probing to test whether encoder-free and discrete-token MLLMs progressively organize perceptual inputs into linearly decodable concepts, in contrast to encoder-full MLLMs whose Transformer inputs have already passed through a dedicated encoder.

\begin{figure}[t]
\centering
\includegraphics[width=0.92\linewidth]{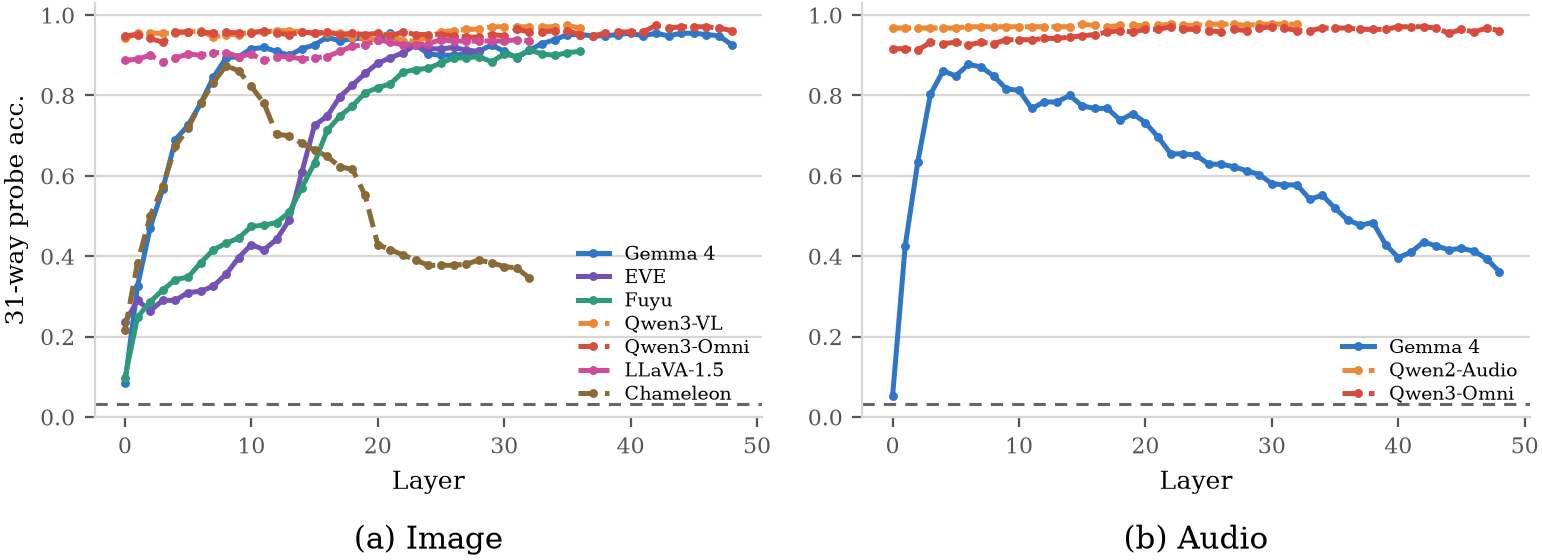}
\caption{Layerwise $|\mathcal{C}|$-way linear-probe accuracy. Encoder-free and discrete-token MLLMs develop concept decodability inside the Transformer, whereas encoder-full controls start with highly decodable representations at layer~0.}
\label{fig:probe}
\end{figure}

\paragraph{Experimental setup.}
We ask when concept identity becomes linearly decodable. For $(K_\text{V},K_\text{A},c)\in\mathcal{D}$ and modality $m\in\{\text{V},\text{A}\}$, feeding sample $K_m$ to a model produces a set of modality-token positions $\mathcal{T}_m$.\footnote{For simplicity, we omit the modality subscript $m$ when there is no ambiguity.} We index these positions by $t\in\mathcal{T}_m$, denote their hidden states at depth (or layer) $l$ by $h^{(l)}_{t}\in\mathbb{R}^{d}$, and represent the image or audio sample by mean-pooling its modality-token hidden states:
\begin{equation}
z^{(l)}=\frac{1}{|\mathcal{T}_m|} \sum_{t \in \mathcal{T}_m} h^{(l)}_{t}.
\label{eq:probe-pooling}
\end{equation}
For each $l$, we independently fit an $\ell_2$-regularized linear concept classifier $f_l(z^{(l)})=W_lz^{(l)}+b_l\in\mathbb{R}^{|\mathcal{C}|}$ using the representations and concept labels in $\mathcal{D}$ and evaluate classification accuracy on $\mathcal{D}_\text{H}$.

\paragraph{Results.}
The layer-wise accuracy is shown in Fig.~\ref{fig:probe}. In the encoder-free MLLMs, semantic information becomes increasingly linearly decodable with depth. For \gemma{} on images, accuracy rises from $8.4\%$ at the projection output at layer 0 to a peak of $95.5\%$ at layer 21, and EVE ($23.5\%\!\to\!92.6\%$ at layer 0 to 23) and Fuyu ($9.7\%\!\to\!91.3\%$ at layer 0 to 32) show similar trends. In contrast, the encoder-full Qwen3-VL and LLaVA-1.5 already exhibit highly decodable representations at layer 0, with Qwen3-VL reaching $94.2\%$. Their Transformer inputs have already been organized by a dedicated modality encoder.
The discrete-token Chameleon also builds comparable decodability only after a substantial stretch of Transformer computation. Its decodability is low at layer 0, climbs to a peak at layer 8, and then \emph{declines} in later layers.

The audio pathway of the encoder-free MLLM; \gemma{} exhibits an even more pronounced depth-dependent pattern. Probe accuracy climbs from $5.2\%$ at layer 0 to $87.7\%$ at layer 6, but then declines to $36.1\%$ by the final layer. In contrast, the encoder-full MLLM; Qwen2-Audio is $96.8\%$ decodable at layer 0 and stays above $97\%$ throughout. \gemma{} thus forms a strongly decodable audio representation early but does not maintain that level of decodability downstream.

\begin{findingbox}{Finding 1}
Encoder-free and discrete-token MLLMs start with weak linear decodability and develop it over depth, whereas encoder-full MLLMs are already highly decodable at the Transformer input.
\end{findingbox}

\subsection{Early Layers Exhibit Encoder-like Representations}
\label{sec:cka}

We test whether hidden states in discrete-token and encoder-free MLLMs share geometric structure with representations from a dedicated single-modal encoder. Encoder-full MLLMs provide controls whose Transformer inputs have already been organized by such an encoder.

\paragraph{Experimental setup.}
We compare each Transformer layer with DINOv2~\cite{oquab2024dinov2} for vision and Audio Spectrogram Transformer (AST)~\cite{ast} for audio using linear CKA \cite{kornblith2019cka}. For each $(K_\text{V},K_\text{A},c)\in\mathcal{D}_\text{H}$ and modality $m\in\{\text{V},\text{A}\}$, we use Eq.~\eqref{eq:probe-pooling} to form a sample-level perceptual representation from $K_m$. We define $X_l\in\mathbb{R}^{|\mathcal{D}_\text{H}| \times d}$ by vertically stacking and mean-centering these representations at depth $l$. We similarly define $Y_{l'}\in\mathbb{R}^{|\mathcal{D}_\text{H}| \times d'}$ for reference-encoder depth $l'$, where $d'$ is the dimensionality of the reference representation. CKA compares how the two models organize whole samples without requiring their token grids or feature dimensions to match, which is given by
\begin{equation}
\operatorname{CKA}(X_l,Y_{l'})
=\frac{\lVert X_l^\top Y_{l'}\rVert_\text{F}^2}
{\lVert X_l^\top X_l\rVert_\text{F}\,
 \lVert Y_{l'}^\top Y_{l'}\rVert_\text{F}}.
\label{eq:cka}
\end{equation}
We use the debiased variant of the linear CKA estimator \cite{murphy2024biasedcka}. We use the same image-level pooling for all six vision-capable models.
We do the same for the audio modality using AST as our reference encoder.

Even if CKA against an early reference-encoder layer establishes geometric similarity, it does not by itself show that the model processes low-level properties such as edges or textures. We therefore complement CKA with a direct sensitivity test in \gemma{} only for visual input. If its early states depend on low-level visual information carried by high spatial frequencies, removing high-frequency information should substantially change their geometry. For each RGB channel of a clean image $I$, we apply a two-dimensional Fourier transform, retain the centered circular region whose radius is $12.5\%$ of the maximum frequency radius, and set all coefficients outside this region to zero. We then apply the inverse transform to obtain the low-pass image $\tilde{I}$ and measure
\begin{equation}
S_l=1-\operatorname{CKA}(X_l,\tilde{X}_l),
\label{eq:lowlevel-sensitivity}
\end{equation}
where $\tilde{X}_l$ is the column-centered representations of $\tilde{I}$ computed in the same way as $X_l$ over $60$ images randomly sampled from $\mathcal{D}_\text{H}$. Larger $S_l$ means that removing high-frequency visual information more strongly reorganizes the representation geometry.

\begin{figure}[t]
\centering
\includegraphics[width=\linewidth]{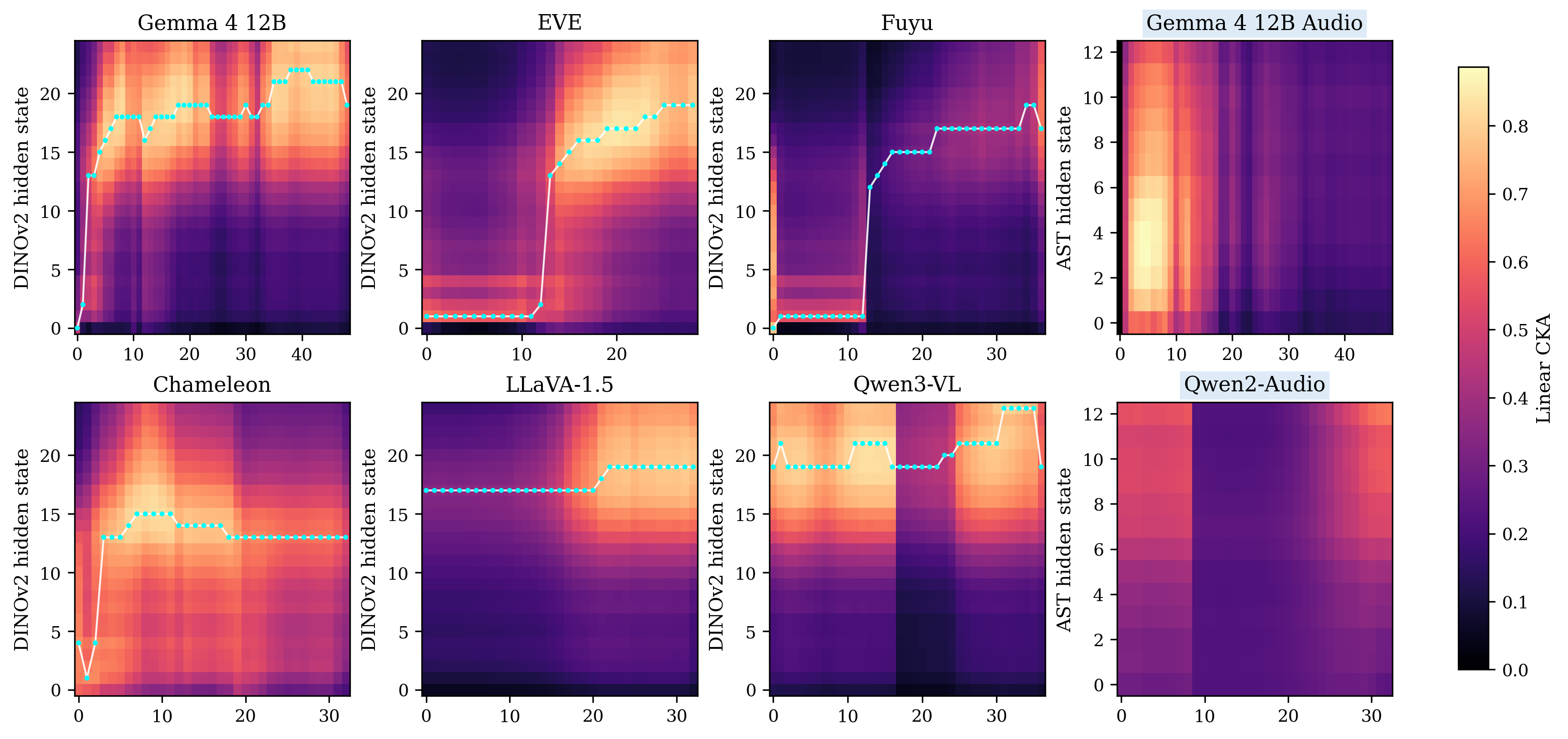}
\caption{Image-level CKA. The horizontal axis is the target model's hidden state; the vertical
axis is DINOv2 hidden state for vision and AST hidden state for audio. The first three columns show vision against DINOv2; the last column shows audio against AST. Cyan markers trace the DINOv2 layer with the highest CKA for each MLLM layer.}
\label{fig:cka_grid}
\end{figure}

\paragraph{Results.}
Figure~\ref{fig:cka_grid} compares six vision-capable models spanning the three
families shown in Figure~\ref{fig:architecture}. The encoder-free MLLMs are
\gemma{}, EVE, and Fuyu, the discrete-token MLLM is Chameleon, and the encoder-full MLLMs are LLaVA-1.5 and Qwen3-VL. The cyan markers, which trace the best-matching DINOv2 layer, start in the lower-left for the encoder-free and discrete-token MLLMs and then move upward. Their early layers thus organize images most like the early layers of a dedicated encoder before resembling its later layers. In contrast, the markers of the encoder-full models stay in the upper region from layer 0. For example, the first layers best match DINOv2 layer 0 to 2 in \gemma{}, EVE, and Fuyu, whereas layer 0 of Qwen3-VL best matches DINOv2 layer 19 (CKA $0.74$). The inputs of encoder-full MLLMs are therefore already organized like late encoder layers.

The audio column shows a related contrast. \gemma{}, an encoder-free MLLM, begins nearly unaligned with AST at layer 0 (CKA $0.00$), develops strong AST-like geometry (CKA $0.89$ at layers 4 and 5), and declines to $0.23$ by layer 48. In contrast, the encoder-full model Qwen2-Audio starts at $0.55$ and remains high, reaching $0.64$ at its final state. Thus AST-like geometry develops inside \gemma{}, whereas Qwen2-Audio receives features already shaped by its audio encoder.

The perturbation test clarifies what the early layers of \gemma{} process. At layer 1, high-frequency removal reduces clean-to-perturbed CKA to $0.48$ ($S_1=0.52$). The sensitivity falls to $0.27$ at layer 48. Thus the early representation is especially sensitive to fine-scale image structure. Together with its image-level similarity to early DINOv2 layers, this provides evidence for low-level visual processing in \gemma{}.

\begin{findingbox}{Finding 2}
The early layers of models given patches or discrete image tokens best match early DINOv2 layers, whereas models given continuous encoder features best match late DINOv2 layers from their first layer. In \gemma{}, direct sensitivity to high-frequency removal further shows that the early layers process low-level visual structure. These similarities describe how the representations are organized.
\end{findingbox}

\subsection{The Virtual Encoder's Output is Causally Read Out at Model-Specific Depth}
\label{sec:causal}

In an encoder-full MLLM, perceptual information reaches the multimodal Transformer only after the dedicated encoder has completed its computation. In discrete-token and encoder-free MLLMs, perceptual tokens enter the multimodal Transformer directly and can transfer information to other token positions from the first block. We perturb modality-token states at individual depths to determine when later computation ceases to depend on those states.

\paragraph{Experimental setup.}
Controlled corruption of internal activations has been used to test whether particular states causally contribute to a model prediction~\cite{meng2022locating}. We apply this intervention principle across depth to determine when the output for sample $K_m$ ceases to depend on its modality-token states. Specifically, we perturb its modality-token positions $t \in \mathcal{T}_m$ after Transformer layer $l$ as follows:
\begin{equation}
\tilde h^{(l)}_{t}=
\begin{cases} 
h^{(l)}_{t} + \epsilon^{(l)}_{t}, & \text{if } t \in \mathcal{T}_m, \\ 
h^{(l)}_{t}, & \text{if } t \notin \mathcal{T}_m, 
\end{cases}
\qquad
\epsilon^{(l)}_{t}\sim\mathcal{N}(0,\sigma_{l}^{2}\mathbb{I}_d),
\label{eq:causal-corruption}
\end{equation}
where $\mathbb{I}_d$ is the $d$-dimensional identity matrix. The states of all other token positions and layers remain unchanged. We sweep one corrupted block at a time over the samples in $\mathcal{D}_\text{C}$. For \gemma{}, we use $\sigma_{l}=500$ for both the image and audio, sweeping all layers. For the images, we additionally repeat all layers with $\sigma\in\{250,1000\}$ to test sensitivity to corruption strength. For Chameleon, whose hidden states' scale changes substantially with depth, we use $\sigma_{l}=20\times\operatorname{RMS}(\{h^{(l)}_{t}\}_{t=1}^{T_m})$, where RMS is
taken across all modality-token positions and hidden dimensions.

We relabel the samples in $\mathcal{D}_\text{C}$ as \emph{animated} or \emph{non-animated} based on their concept labels and use the prompt ``\texttt{Is this alive? Answer by either yes or no.}''
Besides accuracy, we measure the change in the \texttt{yes}/\texttt{no} logit margin. For the $n$-th selected sample, let $o_v^{(n)} \in \mathbb{R}$ denote the next-token logit assigned to vocabulary item $v$. The margin is given by
\begin{equation}
\delta_n=o_{\texttt{yes}}^{(n)}-o_{\texttt{no}}^{(n)}.
\end{equation}
We compute the same margin after perturbing the hidden states at depth $l$, denoted by $\tilde{\delta}^{(l)}_n$. We define the mean absolute difference between clean and perturbed logit margins as
\begin{equation}
    \Delta_l=\frac{1}{|\mathcal{D}_\text{C}|}\sum_{n=1}^{|\mathcal{D}_\text{C}|}|\tilde{\delta}^{(l)}_n-\delta_n|,
\label{eq:causal-margin}
\end{equation}
A large $\Delta_l$ or an accuracy drop means the answer still depends on the modality-token residual at layer $l$; a near-zero effect means that information has already been read into other positions.

\begin{figure}[t]
\centering
\includegraphics[width=\linewidth]{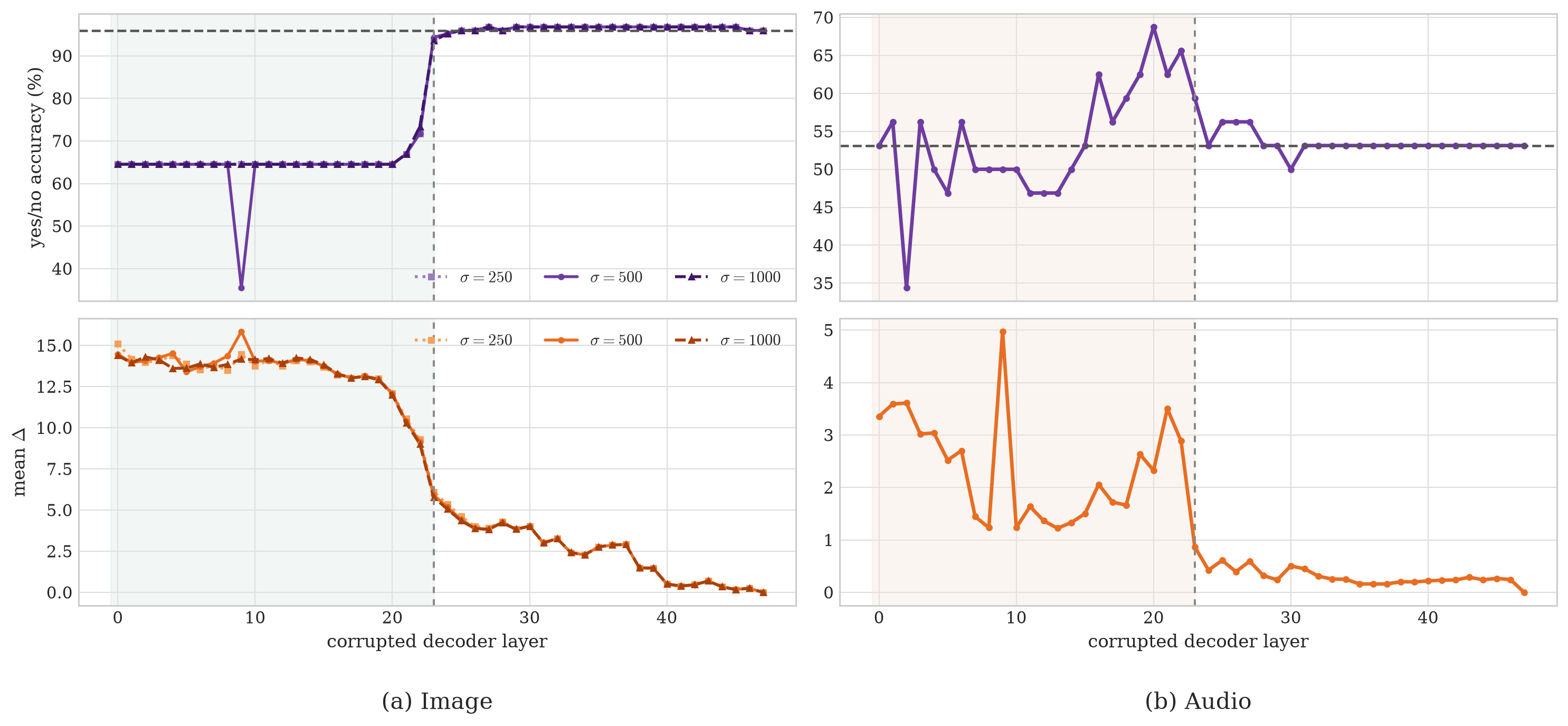}
\caption{Causal readout of modality information measured by single-layer corruption of the modality-token residual across depth. The top row shows classification accuracy, and the bottom row shows the mean absolute shift in the \texttt{yes}--\texttt{no} logit margin. For the image modality (left), accuracy collapses to the majority-class floor when the modality-token residual is corrupted at any layer up to the readout boundary (near layer 23; dashed vertical line), but fully recovers when corruption is applied beyond this boundary. The three image curves, corresponding to $\sigma\in\{250,500,1000\}$, all identify the same transition. The audio modality (right; $\sigma=500$) shows a similar boundary, although the effect of corruption is weaker.}
\label{fig:causal}
\end{figure}

\paragraph{Results.}
Figure~\ref{fig:causal} shows the accuracy and $\Delta_l$ for perturbation at layer $l$. For the image modality, corrupting the image tokens at \emph{any} layer up to layer 22 collapses accuracy to the majority-class floor ($64.5\%$, mean of $\Delta$ logit margin over layers 0 to 22 is $13.42$, $13.50$, and $13.38$ for $\sigma=250$, $500$, and $1000$, respectively), whereas corruption from layer 24 onward leaves the prediction essentially unchanged (accuracy back to $\ge95\%$, margin shift decaying to zero). The image representation carried by the image tokens is thus causally necessary only up to a \emph{readout boundary} near layer 23, close to the depth at which the probe peaks (layer 21). This boundary is insensitive to a four-fold change in corruption strength. For $\sigma=250$, $500$, and $1000$, accuracy is at or below the $64.5\%$ majority floor through layer 20, recovers to
$93.5$--$94.4\%$ at layer 23, and reaches at least $95.2\%$ at layer 24. The margin-shift curves also nearly coincide. The audio pathway shows the same late boundary (around layer 22 to 24) but a far
weaker effect, consistent with an audio representation that is only weakly usable to begin with.

\begin{figure}[t]
\centering
\includegraphics[width=\linewidth]{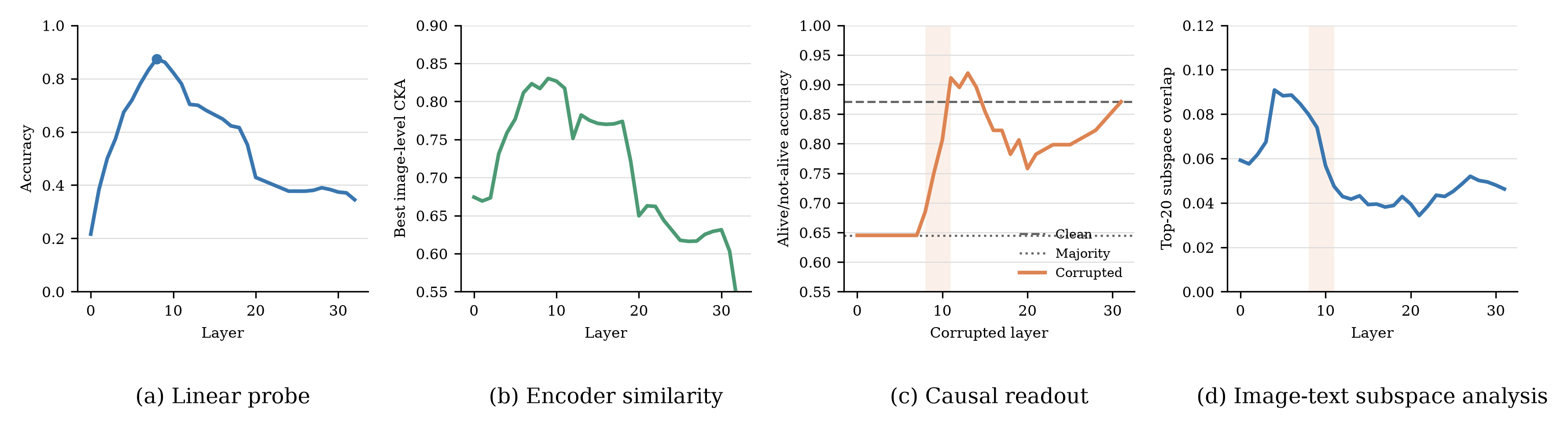}
\caption{Converging evidence in Chameleon. (a) Concept decodability peaks at hidden state layer 8. (b) Image-level CKA against DINOv2 is high around the same early-to-middle transition. (c) Corrupting the VQ-image-token residual collapses the alive/not-alive task to its majority floor through layer 7 and recovers across layers 8 to 11 (shaded). (d) Image-text subspace overlap follows a different trajectory from \gemma{}, peaking earlier and decreasing across the causal readout window.}
\label{fig:chameleon-evidence}
\end{figure}

We repeat the image token intervention in Chameleon, whose Transformers receive $1,024$ discrete image tokens. On the held-out image set $\mathcal{D}_\text{C}$, clean classification accuracy is $87.1\%$, and the majority baseline is $64.5\%$. Corrupting layers 0 to 7 reduces accuracy exactly to that floor. It then recovers through a compact transition, reaching $68.5\%$ at layer 8, $75.0\%$ at layer 9, $80.6\%$ at layer 10, and $91.1\%$ at layer 11. (Figure~\ref{fig:chameleon-evidence}c); the absolute logit-margin effect decays
toward zero over the remaining depth. Thus, the discrete-token model causally consumes its image-token representation around layers 8 to 11, close to its probe maximum at layer 8 (Section \ref{sec:probe}) and a region of high image-level encoder similarity (Section \ref{sec:cka}).

Together, these interventions localize \emph{when} the Virtual Encoder's output is consumed near the midpoint of \gemma{} and substantially earlier in Chameleon. They establish necessity before the respective readout transitions. They do not pinpoint a single encoding layer, since corrupting any layer in the network propagates, and a sufficiency test via activation patching remains for future work.

\begin{findingbox}{Finding 3}
The output depends strongly on image-token residuals through layer 22 in \gemma{}, but Chameleon begins recovering across layers 8 to 11. The readout depth is therefore an empirical property of each model.
\end{findingbox}

\subsection{An Early-Layer Virtual Encoder Regime}
\label{sec:regime}

Taken together, the probe, representational similarity, and causal readout identify an early-to-middle regime in \gemma{} that behaves as an implicit modality encoder. We refer to this computational role as the Virtual Encoder. Related probe and representational signatures appear in other models without continuous encoder features, while Chameleon exhibits a similar but earlier causal readout transition. A Virtual Encoder is a functional description whose presence and depth must be established empirically.

\section{Characterizing the Virtual Encoder through Modality Subspace Allocation}
\label{sec:subspace}

We next ask whether the different downstream behaviors of vision and audio are reflected in how their representations occupy the shared residual stream. We focus on Gemma 4 12B, where image, text, and audio tokens are processed within the same residual stream by the same set of transformer weights. This shared $d$-dimensional space allows us to directly compare the modality-specific subspaces formed at each depth and to ask where the Virtual Encoder places its output relative to the language subspace. In particular, because all modalities share a common coordinate system, their subspace geometry can be quantified using principal angles, unlike the cross-model comparison in Section \ref{sec:cka}, where we instead relied on the angle-invariant CKA measure.

\paragraph{Experimental setup.}
For modality $m \in \{\text{V}, \text{A}, \text{T}\}$ at depth $l$ of a target model, where $\text{T}$ means the text modality, we stack the pooled and centered representations $X_m^{(l)}\in\mathbb{R}^{|\mathcal{D}_\text{C}|\times d}$ as in Eq.~\eqref{eq:probe-pooling}. We use the samples in $\mathcal{D}_\text{C}$ and write a description for each sample. For textual representation, we mean-pool over all textual tokens. Let $U_m^{(l)}\in\mathbb{R}^{d\times k}$ contain the top $k$ principal component directions of $X_m^{(l)}$. Following the classical principal-angle construction of Bj{\"o}rck and Golub~\cite{bjorck1973numerical}, the $j$-th singular value $\sigma_j$ of $U_m^{(l)\top}U_{m'}^{(l)}$ for $m' \in \{\text{V}, \text{A}, \text{T}\}$ equal the cosines of the $j$-th principal angle $\theta_j$.
We define their mean as a metric for overlap summary, i.e.,
\begin{equation}
\operatorname{Overlap}_{m,m'}^{(l)}
=\frac{1}{k}\sum_{j=1}^{k}\sigma_j
=\frac{1}{k}\sum_{j=1}^{k}\cos\theta_j.
\label{eq:subspace-overlap}
\end{equation}
Identical subspaces for modalities $m$ and $m'$ give $\operatorname{Overlap}_{m,m'}^{(l)}$ being 1, and orthogonal subspaces gives $0$. To calibrate this scale, we sample $2,000$ pairs of independent random $k$-dimensional subspaces in the same $d$-dimensional residual space and report their mean overlap. We repeat
the analysis for $k\in\{5,10,20,40\}$. Following cross-projected variance analyses \cite{elsayed2016reorganization}, we also report the fraction of variance in modality $m$ captured by the subspace of modality $m'$ as follows.
\begin{equation}
\operatorname{VarCap}_{m\rightarrow m'}^{(l)}
=\frac{\|X_m^{(l)}U_{m'}^{(l)}\|_\text{F}^2}{\|X_m^{(l)}\|_\text{F}^2}.
\label{eq:variance-capture}
\end{equation}

We use the encoder-full Qwen3-Omni as a control. We apply the same image-text analysis to the discrete-token Chameleon using the $|\mathcal{D}_\text{C}|$ descriptions and compare hidden states.

\begin{figure}[t]
\centering
\includegraphics[width=\linewidth]{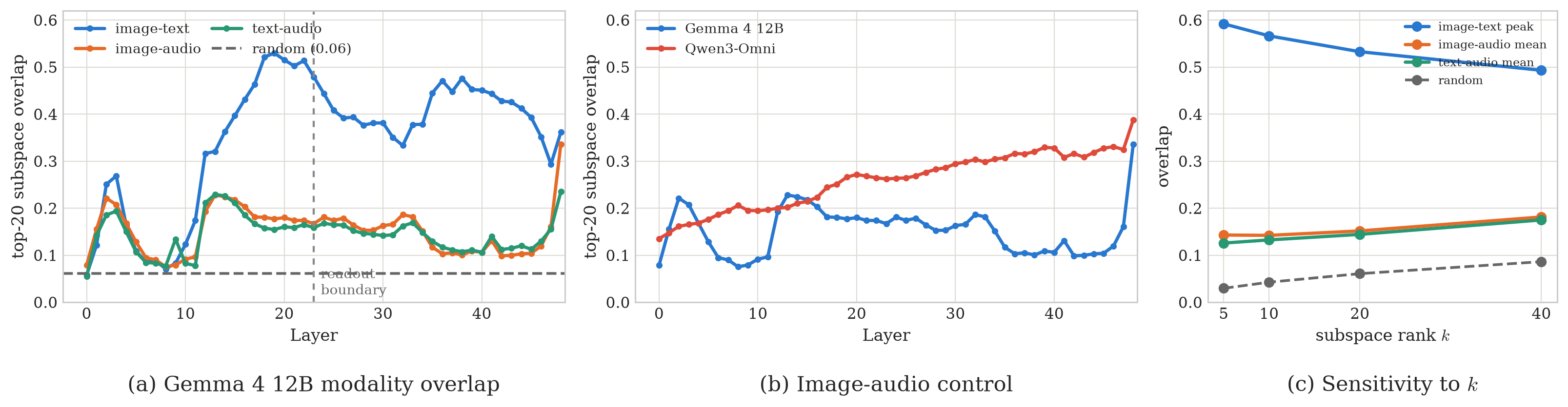}
\caption{Modality subspace allocation in the shared residual stream. (a) Top-$20$ overlap in \gemma{}; the horizontal dashed line is the mean overlap of independent random $20$-dimensional subspaces in the original $d$-dimensional space. Image-text subspace alignment is maximal near the readout transition (vertical dashed line), whereas the audio pairs remain much weaker but above chance. (b) Image-audio overlap for \gemma{} and the encoder-full Qwen3-Omni control. (c) The image-text peak and layer-mean audio overlaps for $k\in\{5,10,20,40\}$, with the corresponding random baseline. }
\label{fig:subspace}
\end{figure}

\paragraph{Results.}
\textbf{The Virtual Encoder rotates vision into the language subspace near the readout boundary.} In \gemma{} (Figure~\ref{fig:subspace}a), the image-text subspace overlap climbs from near-orthogonal at layer 0 ($\operatorname{Overlap}_{\text{V},\text{T}}^{(0)} = 0.06$) to a broad high-overlap band at layers 18 to 22, peaking at layer 19 ($0.53$; $34\%$ of the image subspace's variance lies inside the text subspace), and then falls away again. This band contains the probe peak at layer 21 and lies immediately before the causal readout transition near layer 23 (Section~\ref{sec:causal}). The image subspace is therefore most aligned with the model's language subspace around the depth where the prediction becomes independent of the image-token residual. The image-audio and text-audio overlaps are much weaker. At $k=20$, their layer means are $0.15$ and $0.14$, respectively, which are closer to independence (the random baseline's overlap is $0.061$). They therefore share some structure and should not be
described as strictly orthogonal, but remain far below the image-text peak of $0.53$. This separation is robust to PCA rank as shown in Figure~\ref{fig:subspace}c). Thus, the audio modality does not undergo the pronounced, readout-aligned rotation toward the language subspace seen for vision.

\textbf{The encoder-full control exhibits a different subspace trajectory.}
In the encoder-full control (Figure~\ref{fig:subspace}b), the image-audio overlap rises \emph{monotonically} with depth, with no comparable peak. This trajectory is consistent with progressive cross-modal mixing and does not exhibit the transient geometry observed in \gemma{}.

\textbf{The computational role does not require one universal routing geometry.} 
Chameleon provides a useful counterpoint (Figure~\ref{fig:chameleon-evidence}d). Its image-text overlap rises from $0.059$ at layer 0 to a modest early peak of $0.091$ at layer 4, then decreases across the
causal readout transition to $0.047$ at layer 11. It therefore does not reproduce \gemma{}'s readout-aligned rotation. Thus constructing and consuming a perceptual representation inside the Transformer also occurs in a discrete-token MLLM, while the route taken through the residual stream is architecture dependent. In \gemma{}, modalities occupy only weakly overlapping subspaces, with vision rotating from an image-private subspace toward the language subspace around readout.

Taken together, Sections~\ref{sec:virtual-encoder} and~\ref{sec:subspace} separate three properties of the Virtual Encoder. They show where a perceptual representation is formed, when it is causally consumed, and how it is routed. Gemma and Chameleon support the first two properties, whereas their subspace trajectories differ. A Virtual Encoder offers the computational role of internally constructing a usable perceptual representation before readout, rather than by a particular layer interval or routing geometry.

\section{Discussion}

The Virtual Encoder perspective separates when perceptual representations are formed, when they are read out from modality tokens, and how they are routed afterward. Its strength and depth may depend on how much perceptual processing remains after fusion, placing architectures on a continuum rather than in a fixed encoder-free category. This view makes readout depth and modality-to-language alignment potential design variables. Controlling them may help locate information loss, improve fusion, and preserve perceptual evidence for downstream reasoning.

Our evidence nevertheless leaves important gaps. The full image-text-audio subspace analysis is limited to \gemma{}, while Chameleon provides only an image-text comparison. The language subspace is operationally defined by PCA of text representations. We do not intervene directly on subspace rotation, and audio's weak language alignment is only correlated with its loss of decodability. Controlled front-end comparisons, direct routing interventions, and broader cross-architecture evaluation are therefore needed to establish when Virtual Encoders arise and whether their geometry causally affects model behavior.

\section{Conclusion}

We identify the Virtual Encoder as a regime in which an MLLM internally forms a usable perceptual representation before causal readout. Its depth and routing vary across architectures.
Together, these findings provide a concrete way to identify and study encoder-like computation inside encoder-free multimodal models, rather than assuming it from the architecture alone.

\subsubsection*{Acknowledgements.}
This work was partly supported by JSPS KAKENHI Grant Number JP23H00497 and JST FOREST Grant Number JPMJFR216O.

\bibliographystyle{splncs04}
\bibliography{main}

@inproceedings{lin2023crossmodal,
  title={Multimodality Helps Unimodality: Cross-Modal Few-Shot Learning with Multimodal Models},
  author={Lin, Zhiqiu and Yu, Samuel and Kuang, Zhiyi and Pathak, Deepak and Ramanan, Deva},
  booktitle={Proceedings of the IEEE/CVF Conference on Computer Vision and Pattern Recognition (CVPR)},
  year={2023}
}

@inproceedings{piczak2015esc50,
  title={{ESC}: Dataset for Environmental Sound Classification},
  author={Piczak, Karol J.},
  booktitle={Proceedings of the 23rd ACM International Conference on Multimedia},
  year={2015}
}

@inproceedings{deng2009imagenet,
  title={{ImageNet}: A Large-Scale Hierarchical Image Database},
  author={Deng, Jia and Dong, Wei and Socher, Richard and Li, Li-Jia and Li, Kai and Fei-Fei, Li},
  booktitle={IEEE Conference on Computer Vision and Pattern Recognition (CVPR)},
  year={2009}
}

@inproceedings{radford2021clip,
  title={Learning Transferable Visual Models From Natural Language Supervision},
  author={Radford, Alec and Kim, Jong Wook and Hallacy, Chris and Ramesh, Aditya and others},
  booktitle={International Conference on Machine Learning (ICML)}, year={2021}}

@inproceedings{zhai2023siglip,
  title={Sigmoid Loss for Language Image Pre-Training},
  author={Zhai, Xiaohua and Mustafa, Basil and Kolesnikov, Alexander and Beyer, Lucas},
  booktitle={IEEE/CVF International Conference on Computer Vision (ICCV)}, year={2023}}

@inproceedings{alayrac2022flamingo,
  title={Flamingo: A Visual Language Model for Few-Shot Learning},
  author={Alayrac, Jean-Baptiste and others},
  booktitle={Advances in Neural Information Processing Systems (NeurIPS)}, year={2022}}

@inproceedings{li2023blip2,
  title={{BLIP-2}: Bootstrapping Language-Image Pre-training with Frozen Image Encoders and Large Language Models},
  author={Li, Junnan and Li, Dongxu and Savarese, Silvio and Hoi, Steven},
  booktitle={International Conference on Machine Learning (ICML)}, year={2023}}

@inproceedings{liu2023llava,
  title={Visual Instruction Tuning},
  author={Liu, Haotian and Li, Chunyuan and Wu, Qingyang and Lee, Yong Jae},
  booktitle={Advances in Neural Information Processing Systems (NeurIPS)}, year={2023}}

@inproceedings{dai2023instructblip,
  title={{InstructBLIP}: Towards General-purpose Vision-Language Models with Instruction Tuning},
  author={Dai, Wenliang and Li, Junnan and Li, Dongxu and others},
  booktitle={Advances in Neural Information Processing Systems (NeurIPS)}, year={2023}}

@article{bai2023qwenvl,
  title={{Qwen-VL}: A Versatile Vision-Language Model for Understanding, Localization, Text Reading, and Beyond},
  author={Bai, Jinze and others}, journal={arXiv preprint arXiv:2308.12966}, year={2023}}

@article{wang2024qwen2vl,
  title={{Qwen2-VL}: Enhancing Vision-Language Model's Perception of the World at Any Resolution},
  author={Wang, Peng and others}, journal={arXiv preprint arXiv:2409.12191}, year={2024}}

@article{bai2025qwen3vl,
  title={{Qwen3-VL} Technical Report},
  author={Bai, Shuai and others},
  journal={arXiv preprint arXiv:2511.21631}, year={2025}}

@inproceedings{chen2024internvl,
  title={{InternVL}: Scaling up Vision Foundation Models and Aligning for Generic Visual-Linguistic Tasks},
  author={Chen, Zhe and others},
  booktitle={IEEE/CVF Conference on Computer Vision and Pattern Recognition (CVPR)}, year={2024}}

@article{chu2024qwen2audio,
  title={{Qwen2-Audio} Technical Report},
  author={Chu, Yunfei and others}, journal={arXiv preprint arXiv:2407.10759}, year={2024}}

@article{xu2025qwenomni,
  title={{Qwen3-Omni} Technical Report},
  author={Xu, Jin and others},
  journal={arXiv preprint arXiv:2509.17765}, year={2025}}

@misc{gemma4team2025,
  title={Introducing {Gemma~4}},
  author={{Gemma Team, Google DeepMind}},
  year={2026},
  howpublished={Google DeepMind},
  url={https://deepmind.google/models/gemma/gemma-4/}}

@article{bavishi2023fuyu,
  title={Fuyu-8B: A Multimodal Architecture for AI Agents},
  author={Bavishi, Rohan and others}, journal={Adept AI Blog}, year={2023}}

@inproceedings{diao2024eve,
  title={Unveiling Encoder-Free Vision-Language Models},
  author={Diao, Haiwen and Cui, Yufeng and Li, Xiaotong and others},
  booktitle={Advances in Neural Information Processing Systems (NeurIPS)}, year={2024}}

@inproceedings{diao2025evev2,
  title={{EVEv2}: Improved Baselines for Encoder-Free Vision-Language Models},
  author={Diao, Haiwen and others},
  booktitle={IEEE/CVF International Conference on Computer Vision (ICCV)}, year={2025}}

@article{team2024chameleon,
  title={Chameleon: Mixed-Modal Early-Fusion Foundation Models},
  author={{Chameleon Team}}, journal={arXiv preprint arXiv:2405.09818}, year={2024}}

@article{wang2024emu3,
  title={{Emu3}: Next-Token Prediction is All You Need},
  author={Wang, Xinlong and others}, journal={arXiv preprint arXiv:2409.18869}, year={2024}}

@article{luo2024monointernvl,
  title={Mono-{InternVL}: Pushing the Boundaries of Monolithic Multimodal Large Language Models},
  author={Luo, Gen and others}, journal={arXiv preprint arXiv:2410.08202}, year={2024}}

@article{chen2024solo,
  title={A Single Transformer for Scalable Vision-Language Modeling},
  author={Chen, Yangyi and others}, journal={Transactions on Machine Learning Research (TMLR)}, year={2024}}

@article{oquab2024dinov2,
  title={{DINOv2}: Learning Robust Visual Features without Supervision},
  author={Oquab, Maxime and Darcet, Timoth{\'e}e and others},
  journal={Transactions on Machine Learning Research (TMLR)}, year={2024}}

@inproceedings{alain2016probes,
  title={Understanding Intermediate Layers Using Linear Classifier Probes},
  author={Alain, Guillaume and Bengio, Yoshua},
  booktitle={International Conference on Learning Representations (ICLR) Workshop}, year={2017}}

@inproceedings{kornblith2019cka,
  title={Similarity of Neural Network Representations Revisited},
  author={Kornblith, Simon and Norouzi, Mohammad and Lee, Honglak and Hinton, Geoffrey},
  booktitle={International Conference on Machine Learning (ICML)}, year={2019}}

@article{bjorck1973numerical,
  title={Numerical Methods for Computing Angles Between Linear Subspaces},
  author={Bj{\"o}rck, {\AA}ke and Golub, Gene H.},
  journal={Mathematics of Computation},
  volume={27},
  number={123},
  pages={579--594},
  year={1973},
  doi={10.1090/S0025-5718-1973-0348991-3}}

@article{elsayed2016reorganization,
  title={Reorganization between Preparatory and Movement Population Responses in Motor Cortex},
  author={Elsayed, Gamaleldin F. and Lara, Antonio H. and Kaufman, Matthew T. and Churchland, Mark M. and Cunningham, John P.},
  journal={Nature Communications},
  volume={7},
  pages={13239},
  year={2016},
  doi={10.1038/ncomms13239}}

@inproceedings{murphy2024biasedcka,
  title={Correcting Biased Centered Kernel Alignment Measures in Biological and Artificial Neural Networks},
  author={Murphy, Alex and Zylberberg, Joel and Fyshe, Alona},
  booktitle={ICLR Workshop on Representational Alignment},
  year={2024}}

@inproceedings{raghu2017svcca,
  title={{SVCCA}: Singular Vector Canonical Correlation Analysis for Deep Learning Dynamics and Interpretability},
  author={Raghu, Maithra and Gilmer, Justin and Yosinski, Jason and Sohl-Dickstein, Jascha},
  booktitle={Advances in Neural Information Processing Systems (NeurIPS)}, year={2017}}

@inproceedings{morcos2018insights,
  title={Insights on Representational Similarity in Neural Networks with Canonical Correlation},
  author={Morcos, Ari S. and Raghu, Maithra and Bengio, Samy},
  booktitle={Advances in Neural Information Processing Systems (NeurIPS)}, year={2018}}

@inproceedings{park2024linear,
  title={The Linear Representation Hypothesis and the Geometry of Large Language Models},
  author={Park, Kiho and Choe, Yo Joong and Veitch, Victor},
  booktitle={International Conference on Machine Learning (ICML)}, year={2024}}

@inproceedings{liang2022mindthegap,
  title={Mind the Gap: Understanding the Modality Gap in Multi-modal Contrastive Representation Learning},
  author={Liang, Weixin and Zhang, Yuhui and Kwon, Yongchan and Yeung, Serena and Zou, James},
  booktitle={Advances in Neural Information Processing Systems (NeurIPS)}, year={2022}}

@inproceedings{ansuini2019intrinsic,
  title={Intrinsic Dimension of Data Representations in Deep Neural Networks},
  author={Ansuini, Alessio and Laio, Alessandro and Macke, Jakob H. and Zoccolan, Davide},
  booktitle={Advances in Neural Information Processing Systems (NeurIPS)}, year={2019}}

@inproceedings{valeriani2023geometry,
  title={The Geometry of Hidden Representations of Large Transformer Models},
  author={Valeriani, Lucrezia and others},
  booktitle={Advances in Neural Information Processing Systems (NeurIPS)}, year={2023}}

@inproceedings{jain2025elevating,
  title     = {Elevating Visual Perception in Multimodal LLMs with Visual Embedding Distillation},
  author    = {Jain, Jitesh and Yang, Zhengyuan and Shi, Humphrey and Gao, Jianfeng and Yang, Jianwei},
  booktitle = {NeurIPS},
  volume    = {38},
  pages     = {91092--91123},
  year      = {2025}
}

@inproceedings{venhoff2025visual,
  title     = {How Visual Representations Map to Language Feature Space in Multimodal LLMs},
  author    = {Venhoff, Constantin and Khakzar, Ashkan and Joseph, Sonia and Torr, Philip and Nanda, Neel},
  booktitle = {CVPRW},
  year      = {2025}
}

@inproceedings{ nikankin2025same, title={{Same Task, Different Circuits}: Disentangling Modality-Specific Mechanisms in {VLM}s}, author={Yaniv Nikankin and Dana Arad and Yossi Gandelsman and Yonatan Belinkov}, booktitle={NeurIPS}, year={2025} }

@inproceedings{ast,
  author={Yuan Gong and Yu-An Chung and James Glass},
  title={{AST: Audio Spectrogram Transformer}},
  year=2021,
  booktitle={Interspeech},
  pages={571--575},
  doi={10.21437/Interspeech.2021-698}
}

@inproceedings{meng2022locating,
  author={Kevin Meng and David Bau and Alex Andonian and Yonatan Belinkov},
  title={Locating and Editing Factual Associations in {GPT}},
  booktitle={NeurIPS},
  pages={17359--17372},
  year={2022}
}

@inproceedings{neo2025interpreting,
  title     = {Towards Interpreting Visual Information Processing in Vision-Language Models},
  author    = {Neo, Clement and Ong, Luke and Torr, Philip and Geva, Mor and Krueger, David and Barez, Fazl},
  booktitle = {ICLR},
  year      = {2025}
}

@inproceedings{kaduri2025whats,
  title     = {What's in the Image? {A} Deep-Dive into the Vision of Vision Language Models},
  author    = {Kaduri, Omri and Bagon, Shai and Dekel, Tali},
  booktitle = {CVPR},
  year      = {2025}
}

\end{document}